\documentclass[runningheads]{llncs}
\usepackage[T1]{fontenc}
\usepackage{amsmath}
\usepackage{amssymb}
\usepackage{xcolor}
\usepackage{float} 
\usepackage{booktabs}
\usepackage{url} 

\usepackage{multirow}
\usepackage[table,xcdraw]{xcolor}

\newcommand{\best}[1]{\textbf{{#1}}}
\newcommand{\second}[1]{\underline{{#1}}}
\newcommand{\mtitle}[1]{\noindent \textbf{#1.}}
\newcommand{\lcite}[1]{{\textcolor{red}{[cite]}}}
\newcommand{\he}[1]{H\&E}

\definecolor{Gray}{gray}{0.85}
\definecolor{LightCyan}{rgb}{0.88,1,1}

\newcolumntype{a}{>{\columncolor{LightCyan}}c}
\newcolumntype{b}{>{\columncolor{white}}c}

\usepackage{graphicx}
\usepackage{caption}
\begin{document}
\title{MMAP: A Multi-Magnification and Prototype-Aware Architecture for Predicting Spatial Gene Expression}

\titlerunning{MMAP Architecture for Predicting Spatial Gene Expression}

\author{
Hai Dang Nguyen\inst{1}\orcidID{0009-0008-5752-1049}\thanks{Co-first author.} \and
Nguyen Dang Huy Pham\inst{2}\thanks{Co-first author.} \and
The Minh Duc Nguyen\inst{1} \and
Dac Thai Nguyen\inst{1} \and
Hang Thi Nguyen\inst{3}\thanks{Co-corresponding author.} \and
Duong M. Nguyen\inst {4}\thanks{Co-corresponding author.}
}

\authorrunning{Nguyen and Pham et al.}

\institute{Institute for AI Innovation and Societal Impact, Hanoi University of Science and Technology, Hanoi, Vietnam \and
Amsterdam High School for the Gifted, Hanoi, Vietnam \and
Anatomic Pathology Division, Laboratory Department, Vinmec Times City International Hospital, Vinmec Healthcare System, Hanoi, Vietnam \and 
University of Illinois Urbana-Champaign, Illinois, USA\\
\email{nmduongg@illinois.edu}}

\maketitle              % typeset the header of the contribution
\begin{abstract}
Spatial Transcriptomics (ST) enables the measurement of gene expression while preserving spatial information, offering critical insights into tissue architecture and disease pathology. Recent developments have explored the use of hematoxylin and eosin (\he{})-stained whole-slide images (WSIs) to predict transcriptome-wide gene expression profiles through deep neural networks. This task is commonly framed as a regression problem, where each input corresponds to a localized image patch extracted from the WSI.
However, predicting spatial gene expression from histological images remains a challenging problem due to the significant modality gap between visual features and molecular signals. Recent studies have attempted to incorporate both local and global information into predictive models. Nevertheless, existing methods still suffer from two key limitations: (1) insufficient granularity in local feature extraction, and (2) inadequate coverage of global spatial context.
In this work, we propose a novel framework, MMAP (Multi-MAgnification and Prototype-enhanced architecture), that addresses both challenges simultaneously. To enhance local feature granularity, MMAP leverages multi-magnification patch representations that capture fine-grained histological details. To improve global contextual understanding, it learns a set of latent prototype embeddings that serve as compact representations of slide-level information.
Extensive experimental results demonstrate that MMAP consistently outperforms all existing state-of-the-art methods across multiple evaluation metrics, including Mean Absolute Error (MAE), Mean Squared Error (MSE), and Pearson Correlation Coefficient (PCC). 
% Spatial Transcriptomics (ST) enables the measurement of gene expression while retaining spatially resolved data, providing valuable insights into tissue organization and disease diagnosis. Recent advances have witnessed the utilization of whole slide, hematoxylin and eosin (H\&E)-stained images to enable the profiling of transcriptome-wide gene expression. However, predicting spatial gene expression from these images remains a challenging task, mainly due to the vast difference between the two modalities. Latest works have attempted to integrate both local and global information for spatial gene expression prediction; nevertheless, these efforts have been limited in effectiveness, and have yet to achieve comprehensive modeling of the rich, complex spatial dependencies inherent in histopathological images. Here, we propose MMAP, a two-phase deep learning framework that leverages multi-magnification patch representations and global context-aware refinement to infer gene expression from histology images. The experimental results demonstrate that MMAP outperforms other state-of-the-art baseline methods in prediction accuracy. Our findings highlight that, by leveraging both local information and global context, MMAP can effectively learn rich spatial representations and accurately infer spatially resolved gene expressions from histopathological images.

\end{abstract}
\section{Introduction}
% Single cell RNA sequencing (scRNA-seq) allows researchers to study the gene expression profiles of individual cells within a sample, enabling the detection and quantification of mRNA molecules in a biological sample at the cellular level. However, to fully examine and collect data, cells have to be isolated from whole tissues in an intact and viable condition. This cell isolation process for analysis loses spatial context in the process, since tissue structure and cell location data are lost. Thus, spatial transcriptomics, or the spatially resolved transcriptomic study in intact tissues, emerged to address this issue. Spatial information obtained from ST studies can be helpful in defining the cellular phenotype, cell states, and ultimately their functions in the human body, enhancing our understanding of tissue function and pathology. \cite{ref_article6}
\textbf{Background.} Spatial Transcriptomics (ST) quantifies mRNA expression for a defined set of genes across a tissue sample by segmenting it into discrete "spots". Unlike bulk RNA sequencing, which measures gene expression across an entire tissue and misses intra-sample heterogeneity and spatial relationships, or single-cell RNA sequencing (scRNA-seq), which captures cell-level heterogeneity but loses spatial context due to cell isolation, ST methods preserve both molecular and spatial information. Advanced ST techniques~\cite{exseq,smfish,starmap,visium,seqfish,merfish} enable analysis at varying resolutions, from multi-cell tissue segments~\cite{visium} to single cells~\cite{seqfish} or subcellular regions. Despite their scientific potential, ST methods remain costly, requiring specialized expertise, equipment, and reagents. To this end, deep neural networks have emerged as a promising solution to inferring gene expression profiles directly from histological images.

\noindent \textbf{Deep learning-based spatial transcriptomics prediction.} 
%Early approaches typically framed this task as a regression problem, where a model predicts gene expression from individual image patches, often employing convolutional neural networks (CNNs) or transformer-based architectures to learn visual representations. 
% A typical ST workflow for gene expression prediction begins with representing tissue spots from pathology images, followed by designing predictors to evaluate their molecular profiles. 
Early efforts in gene expression prediction commonly framed the task as a regression problem, wherein models were trained to estimate gene expression levels from individual image patches. These approaches typically employed convolutional neural networks (CNNs) or transformer-based architectures to learn visual representations from histopathological inputs. In this context, each tissue spot was encoded using deep features extracted from intermediate layers of pretrained models, such as ResNet18 or ResNet101~\cite{chung2024accurate,xie2024}, which effectively capture both fine-grained morphological cues and higher-order structural patterns. Building upon this foundation, subsequent methods have sought to enhance predictive performance by leveraging these image-derived feature representations for spatial transcriptomics (ST) data. For example, ST-Net \cite{stnet} applies a transfer learning strategy, fine-tuning a DenseNet121 model pretrained on ImageNet to predict gene expression from histology images. THItoGene \cite{thitogene} introduces a more sophisticated architecture based on dynamic convolutional layers and capsule networks to infer RNA-Seq profiles from whole-slide images (WSIs). Extending this line of work, HisToGene~\cite{histogene} employs a Vision Transformer (ViT) to model patch-level correlations across WSIs, enabling gene expression prediction informed by global spatial context.
Despite their promise, these models are fundamentally limited by their narrow focus on isolated patches, ignoring the broader spatial dependencies inherent in whole-slide images. 
Recent studies~\cite{deeppt,tcgn,hist2st} integrate pathology images and spatial information, employing graph neural networks (GNNs) to model complex spot interactions and spatial relationships within tissues.

 % Recent studies~\cite{deeppt},\cite{hist2st},\cite{tcgn} integrate pathology images and spatial information, employing graph neural networks (GNNs) or vision transformers (ViTs) to model complex spot interactions and spatial relationships within tissues. 

 \noindent \textbf{Limitations of existing approaches and our solution.} However, existing methods still fall short of modeling long-range dependencies and often overlook the hierarchical nature of histopathological information. In practice, gene expression in a given region is influenced not only by its immediate surroundings but also by distant tissue context, sometimes requiring analysis at the whole-slide level to detect disease-specific patterns. Meanwhile, current models typically extract local features only at the patch level, missing fine-grained morphological signals that often become apparent only under higher magnifications (e.g. 20x, 40×). To tackle these challenges, we propose a novel framework that targets both the granularity and spatial scope of information extraction. First, to enhance fine-grained local representation, we introduce a multi-magnification strategy using random cropping to generate sub-patches, allowing the model to access high-resolution details. Second, to capture broader spatial context while maintaining scalability, we learn a set of representative prototype embeddings that summarize global tissue-level patterns. These prototypes act as contextual anchors, enabling the model to integrate both local and global information for more accurate gene expression prediction. To realize these ideas, we introduce MMAP, a novel deep neural network that integrates \textbf{M}ulti-\textbf{MA}gnification features and builds a \textbf{P}rototype bank to seamlessly combine local and global information for gene expression prediction, while maintaining computational efficiency. MMAP operates in two key stages. Firstly, it generates spot-level features by leveraging multi-magnification views to capture detailed local context. Next, these features are used to create a prototype bank - a compact set of representative embeddings that encapsulate the entire WSI. This bank serves as a condensed representation of the WSI, providing rich global context without the need to process thousands of image patches. By employing a cross-attention mechanism, MMAP effectively combines multi-magnification features and models interactions between spots within the tissue. This approach enhances prediction accuracy by utilizing both magnification-specific details and spatial relationships, all while minimizing computational overhead. \\
 
\noindent \textbf{Our contributions.} 
The key contributions are summarized as follows:
\begin{itemize}
    \item We present a novel framework for predicting spatial gene expression levels directly from WSIs. Our approach is designed to capture local histological features at multiple levels of granularity while simultaneously leveraging global contextual information to improve spatial relationship modeling. Crucially, the proposed method maintains computational efficiency and scalability, making it well-suited for the analysis of high-resolution WSIs.
    \item To achieve this, we introduce a multi-magnification feature extraction strategy, which allows the model to learn visual representations across different spatial scales. This mechanism is further supported by auxiliary training objectives that guide the extraction of both fine-grained morphological patterns and broader tissue-level structures.
    \item Furthermore, we incorporate a prototype-based spatial modeling component, in which a set of prototype embeddings is learned to represent the most salient and recurring patterns across the entire slide. This prototype bank serves as a compact and informative summary of the global tissue context, enabling the model to reason about long-range spatial dependencies without the computational burden associated with dense pairwise patch interactions. 
    \item We conduct comprehensive experiments to evaluate the performance of MMAP and compare it with existing approaches. Empirical results demonstrate the superiority of our method against state-of-the-arts. 
\end{itemize}

\section{Related Work}
This section delves into existing studies pertinent to our research. From now on, we refer to the term "spot" as a predefined image patch within a WSI where gene expression is quantified, and use them interchangeably.\\

\mtitle{Deep features for WSIs}
Advancements in deep learning have popularized feature extraction from pretrained networks like ResNet101~\cite{Srinidhi2020-yd,yao_whole_2020}, offering a strong foundation for pathology image analysis. Leveraging these features, Shao et al.~\cite{shao2024} propose a graph embedding algorithm driven by tumor microenvironment interactions to represent image patches. Similarly, Chan et al.~\cite{Chan_2023_CVPR} introduce a heterogeneous graph learning approach to capture local structures in whole-slide image (WSI) pathology data.
Recent developments in pathology foundation models are revolutionizing both general and medical AI, enabling the creation of versatile, general-purpose models that can be either frozen or fine-tuned to extract deep features from pathology images. For example, Lenz et al.~\cite{Lenz_2025_CVPR} integrate features from multiple foundation models: {UNI}~\cite{UNI}, {CTransPath}~\cite{cTransPath}, {Virchow2}~\cite{virchow2}, and {H-optimus-0}~\cite{hoptimus} to generate comprehensive slide-level representations. Likewise, Song et al.~\cite{song2024} utilize UNI~\cite{UNI} to develop morphological slide-level representation sets. In addition, recent solutions~\cite{sema,fddm} employ parameter-efficient fine-tuning techniques, such as low-rank adaptation~\cite{hu2021lora}, to tailor general-purpose models for specific pathology tasks. \\

\mtitle{Spatial gene expression prediction from WSIs}
We review key contributions in predicting spatial gene expression from whole-slide images (WSIs). {ST-Net}~\cite{stnet} employs a transfer learning approach, fine-tuning a DenseNet121 model~\cite{densenet}, pretrained on ImageNet, to predict gene expression from histology images. Building on this, {HisToGene}~\cite{thitogene} uses Vision Transformer to capture patch correlations across WSIs, enabling gene expression prediction with global context-aware features. {Hist2ST}~\cite{hist2st} and {TCGN}~\cite{tcgn} advance this further by incorporating neighborhood information through graph convolutional networks~\cite{gnn}, emphasizing inter-spot relationships. In contrast, {EGN}~\cite{egn} adopts exemplar learning to predict gene expression by selecting the most relevant exemplars from a spot within a WSI. However, this approach neglects local context around patches, compromising the balance between local and global feature integration.

\section{Proposed Method}

\begin{figure}[t]
    \centering
    \includegraphics[width=1\textwidth]{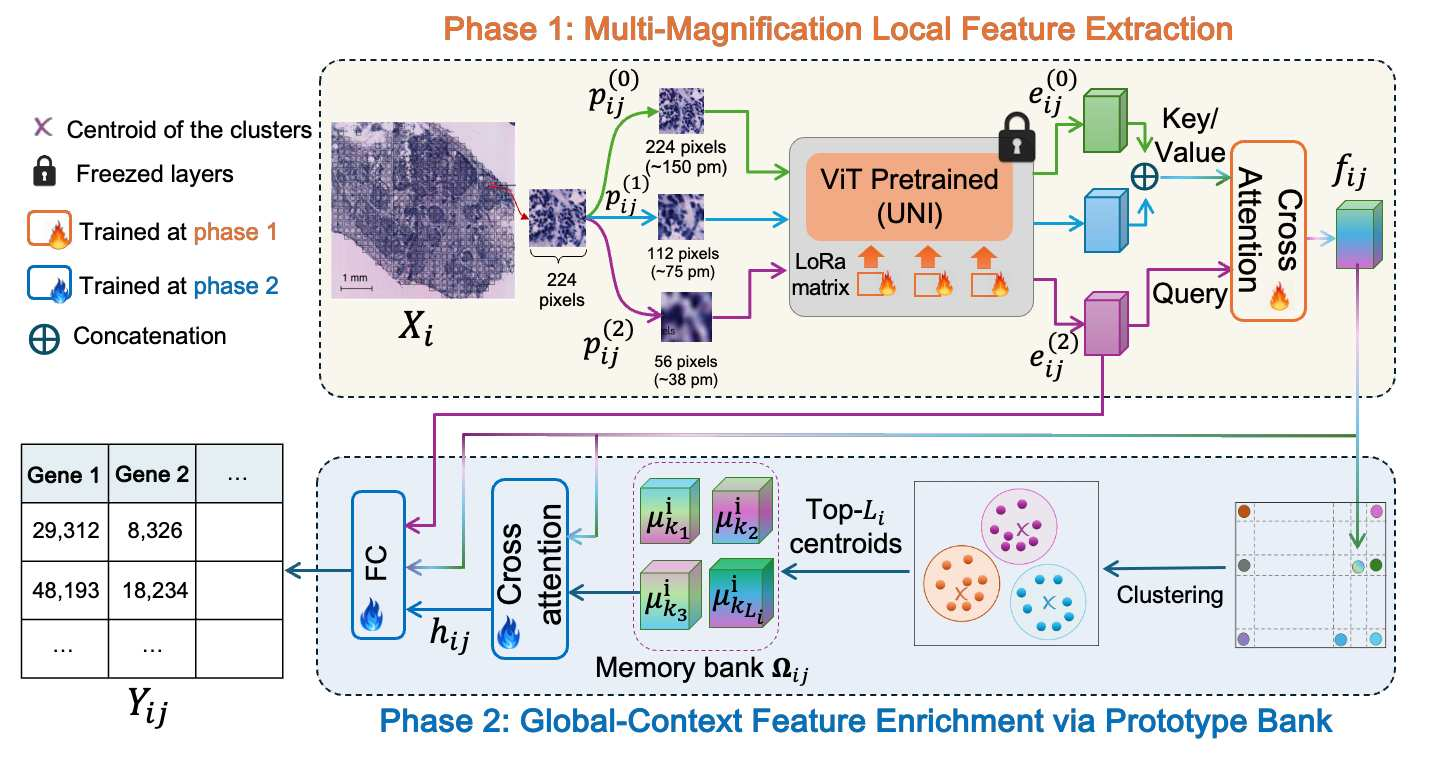}
    \caption{Overview of the proposed framework, comprising two main phases: (1) Local feature extraction, and (2) global context-aware feature enrichment. }
    \label{fig:overview}
\end{figure}

\subsection{Problem Definition and Our Approach}
\subsubsection{Problem Definition.} The goal of spatial gene expression (SGE) prediction is to estimate spatially resolved gene expression profiles directly from Hematoxylin and Eosin (H\&E)-stained WSIs, enabling in-silico molecular profiling without requiring costly spatial transcriptomics assays. Formally, given a WSI $X_i \in {[0, 255]}^{h \times w}$, the tissue section is partitioned into $N_i$ non-overlapping image patches $\{P_{ij}\}_{j=1}^{N_i}$ of fixed size $p \times p$, each centered at the  coordinates $C_{ij}$ corresponding to spatial transcriptomic spot $Y_{ij} \in \mathbb{R}^g$, where $g$ is the number of target genes. The task is to learn a regression model $\mathcal{F}_\theta$ that predicts \begin{equation} \label{eq:prob_def_1}
    \hat{Y}_{ij} = \mathcal{F}_\theta(P_{ij}, C_{ij}),
\end{equation}
such that $\hat{Y}_{ij}$ approximates $Y_{ij}$ as closely as possible.

\subsubsection{Our Solution.}
We introduce a unified framework that integrates multi-magnification local features with global prototypes to enhance gene expression prediction. This reformulates Eq.~\ref{eq:prob_def_1} as follows:
\begin{equation*}
    \mathcal{F}_{\theta}(P{ij}, C_{ij}) = \mathcal{F}^g_{\theta_2} \left( \mathcal{F}^{\ell}_{\theta_1} (P{ij}, C_{ij}) \right),
\end{equation*}
where $\mathcal{F}^{\ell}_{\theta_1}$ extracts multi-magnification local features from patch $P_{ij}$, which are then used by $\mathcal{F}^{g}_{\theta_2}$ to construct a prototype bank for refining local features with global context.

Specifically, our approach operates in two phases. First, the patch-level feature extractor $\mathcal{F}^{\ell}_{\theta_1}$ learns discriminative morphological representations from image patches, encoding them into latent embeddings that capture key histological traits predictive of gene expression. 
In the second phase, a global context enhancement module $\mathcal{F}^{g}_{\theta_2}$ constructs a prototype bank and employs cross-attention to model long-range dependencies across spatially distributed patches. This enriches local representations with global tissue context, enabling effective reasoning about both localized structures and broader WSI architecture. As shown in Fig.~\ref{fig:overview}, this two-phase design balances expressive power and computational efficiency, fully leveraging the richness of whole-slide histopathology images while maintaining scalability and predictive accuracy.

\subsection{Multi-magnification Local Feature Extraction} \label{sec:extraction}

% The first phase focuses on learning a robust patch-level feature extractor $\mathcal{F}_\theta$ that captures informative morphological patterns relevant to gene expression.
% \mtitle{Overview} The goal of this phase is to learn an informative and robust representation for each individual patch. The core idea underpinning our approach is to extract information from the input patch across multiple magnification levels. This design is inspired by the diagnostic workflow of pathologists, who routinely examine histology slides at varying levels of magnification, as each level reveals distinct and complementary biological cues.
% At a high level, for each patch and at every training iteration, we generate magnified views by randomly cropping sub-patches at scales of 1/2 and 1/4 the original patch size, approximating zoom-in operations of $\times 10$ and $\times 20$ relative to the original whole-slide image. The input patch and its corresponding sub-patches are then passed through carefully designed embedding modules that extract salient features specific to each magnification level.
% To guide the model’s focus toward the most informative regions, we incorporate attention layers that emphasize relevant features, and introduce auxiliary loss functions to further enhance the effectiveness of representation learning. In the following sections, we detail the architecture of the embedding modules, the attention mechanism, and the objective functions used during training.
% Figure \ref{fig:phase1} illustrates the details of this phase. 

\begin{figure}[tbh]
    \centering
    \includegraphics[width=1\textwidth]{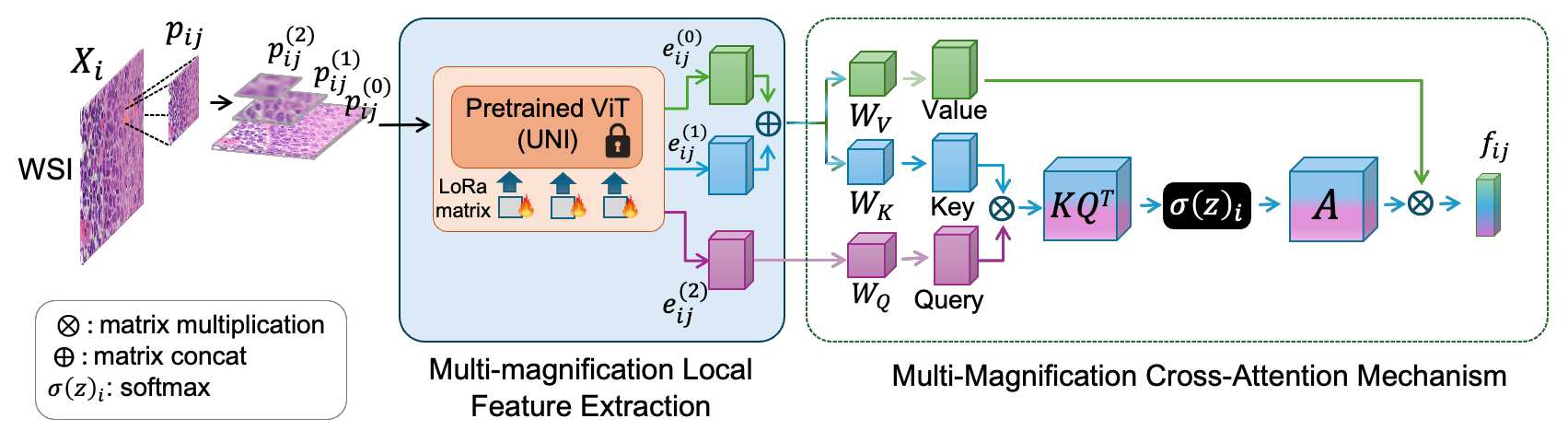}
    \caption{Patch-level feature extraction with Multi-magnification enhancement.}
    \label{fig:phase1}
\end{figure}

In the first phase of {MMAP}, we aim to learn robust patch-level representations by extracting features from histology patches at multiple magnification levels, inspired by pathologists’ diagnostic workflow where varied zooms reveal complementary biological cues. For each patch, we generate sub-patches at 1/2 and 1/4 the original size (approximating $\times 10$ and $\times 20$ magnifications) via random cropping, process them through specialized embedding modules to capture magnification-specific features, and apply attention layers with auxiliary loss functions to focus on informative regions and enhance representation learning, as illustrated in Fig.~\ref{fig:phase1}.

% \subsubsection{Magnification-Aware Feature Extractor}
% For feature extraction, CNN variants~\cite{GRAHAM2019hover, Ho2022, jahanifar2023mitosis} are commonly used to learn \(\mathcal{F}_W^{ct}\) and \(\mathcal{F}_W^{ce}\), but they require extensive datasets, which are scarce in histology due to the need for expert annotations. To address this, we propose fine-tuning UNI \cite{chen2024uni}, a pretrained encoder on a large number of whole-slide images (WSIs), using Low-Rank Adaptation (LoRA) \cite{hu2021lora} to reduce computational costs.

% Let \(d\) and \(k\) be the hidden dimensions of the model parameters, with \(W \in \mathbb{R}^{d \times k}\) as a weight matrix. LoRA updates \(W\) using low-rank decomposition: 
% $W_{t+1} = W_t + \Delta W = W + BA$,
% where \(B \in \mathbb{R}^{d \times r}\) and \(A \in \mathbb{R}^{r \times k}\), with \(r\) as the rank. During training, only \(B\) and \(A\) are updated, keeping \(W\) fixed. This method replaces standard fine-tuning with a low-rank approximation:
% \begin{equation}
%     B \leftarrow B - \alpha \nabla_B \mathcal{L}(BA), \quad A \leftarrow A - \alpha \nabla_A \mathcal{L}(BA),
% \end{equation}
% where \(\alpha\) is the learning rate and \(\mathcal{L}\) is the loss function. In this study, LoRA is applied only to the last \(L\) layers of UNI, as earlier layers capture low-level features.

\subsubsection{Multi-Magnification Cross-Attention Mechanism.} \label{sec:cross-attention}
In this work, we propose a cross-magnification attention mechanism to dynamically model interactions between patches at different resolution levels. For each input patch $P_{ij}$ of size $p \times p$ at the resolution $\times 5$, we generate two additional sub-patches at higher resolutions via random cropping: 
\begin{equation}
P^{(1)}_{ij} \in \mathbb{R}^{\frac{p}{2} \times \frac{p}{2}}, \quad \quad
P^{(2)}_{ij} \in \mathbb{R}^{\frac{p}{4} \times \frac{p}{4}},
\end{equation}
corresponding to approximate magnifications of $\times 10$ and $\times 20$, respectively. All patches are resized back to $p \times p$ for uniform processing, forming a multi-magnification input set:
\begin{equation}
{P}_{ij} = \left\{ \text{resize}(P_{ij}), \text{resize}(P^{(1)}_{ij}), \text{resize}(P^{(2)}_{ij}) \right\}.
\end{equation}
These multi-magnification patches are independently processed by $\mathcal{F}^{\ell}_{\theta_1}$ to obtain magnification-specific embeddings:
\begin{equation}
\mathcal{E}_{ij} = \left\{ \mathbf{e}^{(0)}_{ij}, \mathbf{e}^{(1)}_{ij}, \mathbf{e}^{(2)}_{ij} \right\},
\end{equation}
where $\mathbf{e}^{s}_{ij} = \mathcal{F}^{\ell}_{\theta_1} (P_{ij}^{s})$. Here, $s$ denotes the magnification level of the input patch. In this work, we fine-tune UNI~\cite{UNI}, a foundation model pretrained on a large amount of histological images using LoRA adaption~\cite{hu2021lora}.

Using UNI-based encoder \(\mathcal{F}^{\ell}_{\theta_1}\) on different magnification-specific inputs $\{P_{ij}^s\}$, $\forall s=\{1, 2,3\}$, we generate sequences of $\tau$ vectors \(\{z^{s}_k\}_{k=1}^{\tau} \), respectively, where the first tokens ([CLS] tokens) \(z_0^{s}\) represent the entire sequences. We construct an intermediate sequence \(\{z_0^{0}, z_1^{1}, \dots, z_\tau^{1}, z_1^{2}, \dots, z_\tau^{2}\}\) and apply self-attention to enable the cell [CLS] token \(z_0^{0}\) to interact with features from the higher magnification levels, computing pairwise attention scores to focus on the most relevant information. Intuitively, this multi-magnification attention process aims to enrich representations by looking closer at each sub-region of the original given patch.
The output of this process is a fused representation \(f^{0}_{0}\), which combines \(z_0^{0}\) with magnification-specific features weighted by their attention scores, capturing complex interactions across patches and resolutions in histology images. Henceforth, for simplicity, we drop the subscript \(0\) of the [CLS] token and denote \(f_{ij}\) as the [CLS] vector of the input $P_{ij}$. $f_{ij}$ is then fed into a linear layer to predict gene expressions.

\subsubsection{Training Objectives.}
Annotation inconsistencies and the dropout phenomenon in gene expression typically introduce bias into the training data. To address this and ensure stability during training, we propose a contrastive magnification loss \(\mathcal{L}_{\text{mag}_1}\) to regularize \(f_{ij}\) and reduce dataset bias:
\begin{equation}
    \mathcal{L}_{\text{mag}_1} = 1 - \cos \left( f_{ij}, \mathbf{e}^{(0)}_{ij} \right),
\end{equation}
where $\cos(\cdot)$ denotes the cosine similarity function. This loss encourages the original and magnification-enhanced representations of the same instance to align, ensuring stable training. For gene expression prediction, we employ a simple $\ell_2$ loss function $\mathcal{L}_{\text{ge}_1}$ to train the regression model. Together, the final loss of MMAP in the first stage is as follows:
\begin{equation}
    \mathcal{L} = \mathcal{L}_{\text{ge}_1} + \gamma_1 \mathcal{L}_{\text{mag}_1},
\end{equation}
where $\gamma_1$ controls the impact of regularization losses.

% To improve training stability and enforce consistent embeddings across different magnifications, we apply a contrastive cosine similarity loss between the fused embedding and the highest magnification embedding $\mathbf{e}^{(2)}_{ij}$:

% \begin{equation}
% \mathcal{L}_{\text{contrast}} = 1 - \cos \left( \mathbf{e}^{\text{fused}}_{ij}, \mathbf{e}^{(2)}_{ij} \right).
% \end{equation}

% The total loss for Phase 1 thus combines regression and contrastive objectives:

% \begin{equation}
% \mathcal{L}_{\text{phase1}} = \mathcal{L}_{\text{reg}} + \lambda \mathcal{L}_{\text{contrast}},
% \end{equation}

% where $\lambda$ is a hyperparameter balancing both terms.

% Finally, both the fused embedding and the high-magnification embedding are forwarded into two independent MLP heads for prediction, and their outputs are ensembled by averaging to form the final patch-level gene expression prediction:

% \begin{equation}
% \hat{E}_{ij} = \frac{1}{2} \left( \text{MLP}_1(\mathbf{e}^{\text{fused}}_{ij}) + \text{MLP}_2(\mathbf{e}^{(2)}_{ij}) \right).
% \end{equation}

\subsection{Global-Context Feature Enrichment via Prototype Bank}

After obtaining a robust patch-level feature extractor $\mathcal{F}^{\ell}_{\theta_1}$ from Phase 1, we freeze its parameters and use it as a feature inference backbone for global-context modeling. Specifically, for each WSI ${X}_i$, we process all patches $\{P_{ij}\}$ through phase 1's pipeline to extract two types of intermediate features: (1) the fused embeddings $f_{ij}$ from the multi-magnification cross-attention module, and (2) the corresponding MLP regression outputs $\hat{Y}^{(1)}_{ij}$ from Phase 1.

\subsubsection{Global Prototype Bank Construction.}

This section introduces a novel prototype bank, which serves as a condensed but rich representation set of the WSI. Technically, to capture global tissue context while avoiding full self-attention over all patches (which is computationally prohibitive), we perform $K$-means clustering on the set of fused embeddings $\{\mathbf{e}^{\text{fused}}_{ij}\}$ for each WSI $X_i$ individually:
\begin{equation}
\{\mu^{i}_{k}\}_{k=1}^{K_{i}} = \text{KMeans}\left( \left\{f_{ij}\right\}_{j=1}^{N_i} \right),
\end{equation}
where $\mu^{i}_{k}$ denotes the $k$-th cluster centroid for slide $X_i$, and $K_{i}$ is adaptively selected based on the total number of patches in $X_i$ to balance between over-sparsification and under-representation.
For each patch $P_{ij}$, we retrieve its top-$L_{i}$ most similar cluster centroids according to cosine similarity:
\begin{equation}
{\Omega}_{ij} = \text{TopL}_{L_{i}}\left( \left\{ \mu^{i}_k \right\}, f_{ij} \right),
\end{equation}
where $\Omega_{ij} = \left\{ \mu^{i}_{k_1}, \dots, \mu^{i}_{k_{L_i}} \right\}$ denotes our proposed prototype bank for patch $P_{ij}$. We refer to each embedding in $\Omega_{ij}$ as a global prototype due to its representativeness for the WSI.

We experiment with both fixed and adaptive retrieval strategies for $L_i$. In the adaptive setting, $L_i$ is dynamically adjusted for each WSI based on the total number of clusters $K_i$ to maintain a balance between capturing sufficient global diversity and avoiding overfitting to noisy centroids. As shown in Sec.~\ref{sec:experiment}, we empirically find that adaptive retrieval improves performance in case of highly variable tissue sizes.

\subsubsection{Local-Global Cross-Attention Fusion.}

Given global prototype bank $\Omega_{ij}$, we now aim to fuse information from local and global views to enrich the prediction process. Specifically, we design a global-context aware cross-attention module that integrates local patch features with retrieved global prototypes. Similar to the cross-attention process described in Sec.~\ref{sec:cross-attention}, for each patch $P_{ij}$, we produce a rich representation $h_{ij}$. In this attention process, $f_{ij}$ serves as the queries and $\Omega_{ij}$ as keys and values. Since $h_{ij}$ is now integrated with both local and global information, we linearly project $h_{ij}$ to get another prediction for gene expression. The final prediction of MMAP can be formulated as:
\begin{equation}
    \hat{Y}_{ij} = \left( \text{MLP}_1 (\mathbf{e}_{ij}^{(2)}) + \text{MLP}_2 (f_{ij})  + \text{MLP}_3(h_{ij}) \right) / 3,
\end{equation}
where MLP denotes a linear projection layer.

\subsubsection{Training Objectives.}
To ensure the prototype-enhanced features do not diverge, we similarly adopt the magnification loss \(\mathcal{L}_{mag}\) in Sec.~\ref{sec:extraction} as follows:
\begin{equation}
    \mathcal{L}_{\text{mag}_2} = 1 - \cos \left( f_{ij}, h_{ij} \right).
\end{equation}
A simple $\ell_2$ loss function $\mathcal{L}_{\text{ge}_2}$ to train the regression model, leading the final loss of MMAP in the second stage:
\begin{equation}
    \mathcal{L} = \mathcal{L}_{\text{ge}_2} + \gamma_2 \mathcal{L}_{\text{mag}_2},
\end{equation}
where $\gamma_2$ controls the impact of regularization losses.

\section{Evaluation} \label{sec:experiment}
\subsection{Experimental Settings}

\subsubsection{Dataset.} We conduct experiments on the publicly available HER2-positive breast cancer dataset~\cite{SpatialDeconvHER2}, which includes 36 H\&E-stained WSIs from 8 patients. For each tissue section, histology images, gene expression profiles, and spatial barcode coordinates are provided. We extract fixed-size image patches centered at sequencing spots and retain the top 1,000 highly variable genes (HVGs), further filtering genes expressed in fewer than 1,000 spots across the dataset. This preprocessing results in a total of 9,612 spots and 785 genes. The dataset is split at the slide level: 28 WSIs are used for training, and 8 WSIs with pathologist annotations are reserved for testing. 

\subsubsection{Baseline Methods.} 
We compare our proposed method with four state-of-the-art baseline methods: ST-Net~\cite{stnet}, DeepPT~\cite{deeppt}, HisToGene~\cite{thitogene} and TCGN~\cite{tcgn}. In particular, ST-Net and DeepPT both propose a CNN-based model to learn local features from the whole slide images; HisToGene utilizes a Transformer-based architecture to learn global relationships between patches of a whole slide image; TCGN attempts to combine local and global features while also using Graph Neural Networks (GNNs) to learn inter-spot relationships. These baselines cover different approaches in the field, ensuring the completeness of our experiments.

\subsubsection{Evaluation Metrics.}
We evaluate our method’s performance using three complementary metrics: Pearson Correlation Coefficient (PCC), Mean Squared Error (MSE), and Mean Absolute Error (MAE), all of which are widely adopted in performance testing~\cite{Benchmarking}.
MSE and MAE measure the degree of error between predicted and observed gene expression values by different normalization criteria. Both are used to provide a normalized measure of prediction accuracy. Lower error values indicate better prediction performance.
Meanwhile, PCC measures the linear correlation between predicted and actual gene expression values across spatial locations. A higher PCC value indicates a stronger linear relationship, reflecting the accuracy of spatial gene expression prediction. 

\subsubsection{Implementation Details.} 
All experiments are implemented using the PyTorch framework and trained on a server equipped with a single NVIDIA A100 GPU. We use the Adam optimizer with an initial learning rate of $1 \times 10^{-5}$, no weight decay, and a batch size of 16. The learning rate is scheduled using Cosine Annealing over 50 epochs. Input images are resized to $112 \times 112$ and normalized using the ImageNet mean and standard deviation. Data augmentation includes random horizontal flip, rotation, and color jitter. In our proposed method, the number of clusters $K$ in the K-means step is adaptively selected for each WSI based on its number of extracted patches. Specifically, $K$ is dynamically adjusted within the range $[32, 80]$ to ensure a balanced trade-off between computational efficiency and fine-grained spatial representation. For gene expression values, we apply a $\log(1 + x)$ transformation for normalization prior to training, which helps stabilize variance and reduce the effect of extreme values. 
% Our code will be made publicly available upon publication.

\subsection{Experimental Results} 

\subsubsection{Quantitative Evaluation.}
Table~\ref{tab:comparison} presents a comprehensive comparison between our method (MMAP) and several state-of-the-art baselines on the gene expression prediction task. MMAP consistently outperforms all competitors across key evaluation metrics, demonstrating its superior capability in modeling spatial transcriptomic patterns from histopathology. In particular, MMAP achieves the highest PCC of \best{0.2619}, which is 3.5 times higher than the strongest baseline in this regard, HisToGene (\second{0.0753}). For error-based metrics, MMAP obtains the lowest MSE of \best{1.2439}, representing a 3.8\% reduction compared to the second-best method, DeepPT (\second{1.2822}). Similarly, MMAP achieves a MAE of \best{0.8873}, yielding a 5.2\% improvement over TCGN (\second{0.9396}). These results highlight the effectiveness of our proposed multi-magnification strategy in capturing fine-grained morphological and contextual cues that are predictive of underlying gene expression.

% \begin{table}[H]
% \caption{Comparison of gene expression prediction performance on the test set. \textbf{$\uparrow$}/\textbf{$\downarrow$} means higher/lower values are better. The top and runner-up results are highlighted using \best{bold} and \second{underline}, respectively.}
% \vspace{12pt}
% \centering
% \setlength{\tabcolsep}{4pt}
% \setlength{\doublerulesep}{0.5pt}
% \begin{tabular}{c|ccccc|c}
% \toprule
% Method & \multicolumn{1}{c}{ST-Net} & 
% \multicolumn{1}{c}{HisToGene} & 
% \multicolumn{1}{c}{DeepPT} & \multicolumn{1}{c}{Hist2ST} & \multicolumn{1}{c|}{TCGN} & \begin{tabular}[c]{@{}l@{}}\textbf{MMAP}\\ \textbf{(Ours)}\end{tabular}      \\ \midrule
% PCC $\uparrow$   & 0.0510                     & 0.0753                        & 0.0470                        & {\second{0.2410}}                      & 0.0515                   & {\best{0.2619}} \\ \midrule
% % ARI    & 0.100                      & 0.167                         & 0.010                        & 0.110                       & 0.079                    & {\color[HTML]{FE0000} \textbf{0.080}}  \\ \midrule
% MAE  $\downarrow$  & 0.9580                     & 0.9664                        & 0.9536                        & {\best{0.8858}}                      & 0.9396                   & {\second{0.8873}} \\ \midrule
% MSE  $\downarrow$  & 1.4845                     & 1.4562                        & 1.2822                        & {\second{1.2703}}                      & 1.4858                   & {\best{1.2439}} \\ \bottomrule
% \end{tabular}
% \label{tab:comparison}
% \end{table}

\begin{table}[H]
\caption{Comparison of gene expression prediction performance on the test set. \textbf{$\uparrow$}/\textbf{$\downarrow$} means higher/lower values are better. The top and runner-up results are highlighted using \best{bold} and \second{underline}, respectively.}
\vspace{24pt}
\centering
\setlength{\tabcolsep}{4pt}
\setlength{\doublerulesep}{0.5pt}
\begin{tabular}{c|cccc|c}
\toprule
Method & ST-Net & HisToGene & DeepPT & TCGN & \begin{tabular}[c]{@{}l@{}}\textbf{MMAP}\\ \textbf{(Ours)}\end{tabular}      \\ \midrule
PCC $\uparrow$   & 0.0510  & \second{0.0753}  & 0.0470  & 0.0515  & \best{0.2619} \\ \midrule
MAE  $\downarrow$  & 0.9580  & 0.9664  & 0.9536  & \second{0.9396}  & \best{0.8873} \\ \midrule
MSE  $\downarrow$  & 1.4845  & 1.4562  & \second{1.2822}  & 1.4858  & \best{1.2439} \\ \bottomrule
\end{tabular}
\label{tab:comparison}
\end{table}

% % Please add the following required packages to your document preamble:
% % \usepackage[table,xcdraw]{xcolor}
% % Beamer presentation requires \usepackage{colortbl} instead of \usepackage[table,xcdraw]{xcolor}
% \begin{table}[]
% \centering
% \begin{tabular}{c|cccccc}
% \hline
% Method & \multicolumn{1}{l}{ST-Net} & \multicolumn{1}{l}{HisToGene} & \multicolumn{1}{l}{DeepPT} & \multicolumn{1}{l}{Hist2ST}   & \multicolumn{1}{l}{TCGN} & \multicolumn{1}{l}{MMAP}      \\ \hline
% PCC    & 0.0510                     & 0.0753                        & 0.047                      & {\color[HTML]{3531FF} 0.2410} & 0.0515                   & {\color[HTML]{FE0000} 0.2619} \\ \hline
% ARI    & 0.100                      & {\color[HTML]{FE0000} 0.167}  & 0.0104                     & {\color[HTML]{3531FF} 0.110}  & 0.079                    & {\color[HTML]{333333} 0.080}  \\ \hline
% MAE    & 0.9580                     & 0.9664                        & 0.9536                     & {\color[HTML]{FE0000} 0.8858} & 0.9396                   & {\color[HTML]{3531FF} 0.8873} \\ \hline
% MSE    & 1.4845                     & 1.4562                        & 1.2822                     & {\color[HTML]{3531FF} 1.2703} & 1.4858                   & {\color[HTML]{FE0000} 1.2439} \\ \hline
% \end{tabular}
% \end{table}

\subsubsection{Qualitative Evaluation.}

\begin{figure}[t]
    \centering
    \includegraphics[width=1\textwidth]{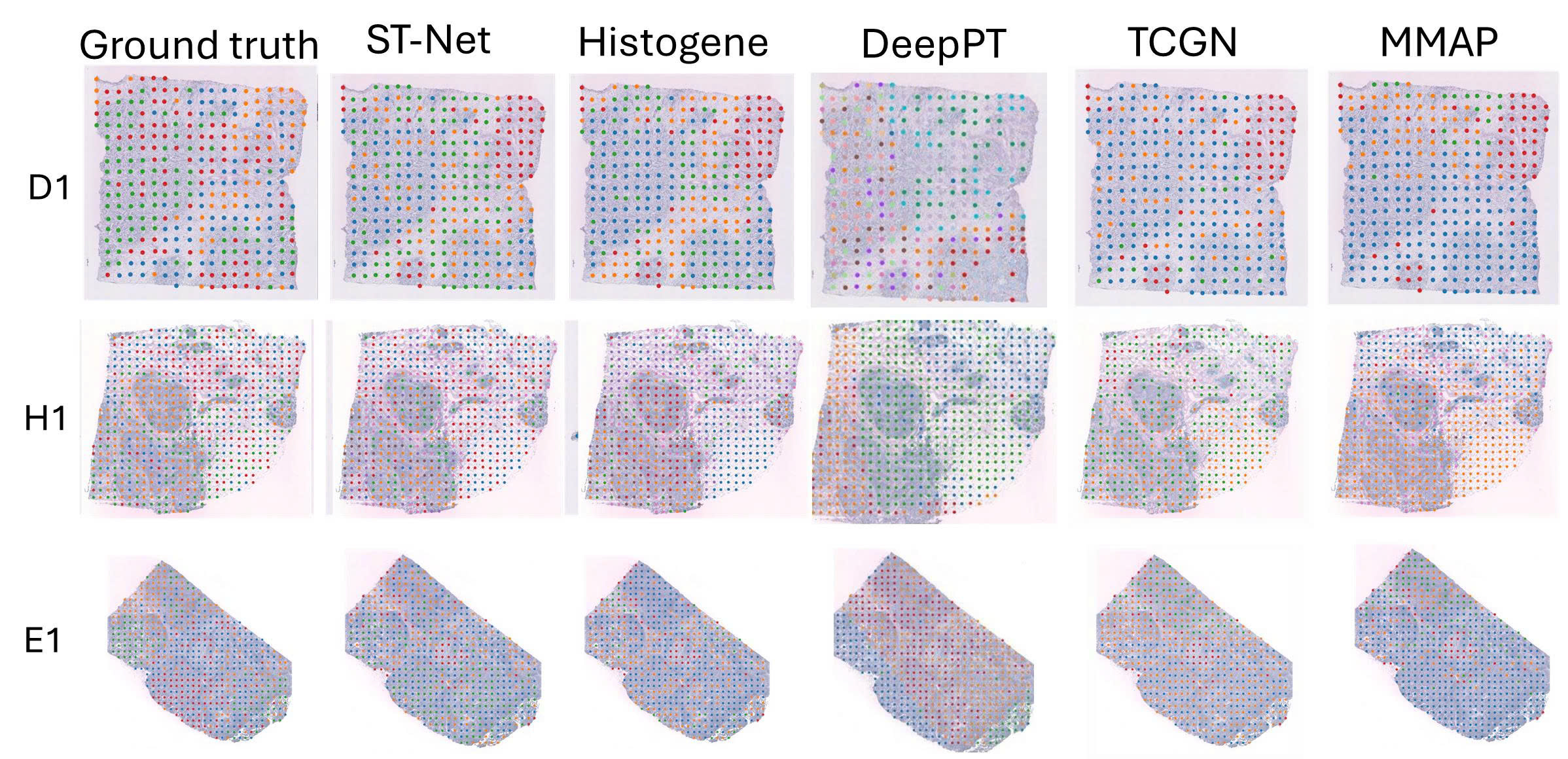}
    \caption{Visualization of gene expression on the HER2+ dataset using prediction results obtained by all methods. The ground truth represents labels from the pathologist annotations. The illustrations for each figure are obtained by performing $K$-means clustering.}
    % , and each method is evaluated in terms of PCC
    \label{fig:test}
\end{figure}

In Figure~\ref{fig:test}, we present qualitative visualizations of predicted gene expression maps on the HER2+ dataset, where K-means clustering is applied to the outputs of each method to support spatial interpretation. Ground-truth annotations derived from pathologist labels are used as references to assess spatial correspondence between predicted expression patterns and annotated tissue regions. MMAP predictions generally follow the structure of the annotated regions and delineate spatial domains with greater visual clarity, particularly in areas characterized by subtle morphological transitions. The clusters produced by MMAP tend to align with histological boundaries and exhibit coherent regional separation. In contrast, clusters generated by several baseline methods often appear spatially fragmented or less congruent with the underlying tissue architecture, resulting in overlapping or dispersed patterns across morphologically distinct regions. These observations highlight the ability of MMAP to produce structured, morphology-consistent predictions that are well-suited for downstream spatial analyses.


\subsection{Ablation Study}

% \begin{figure}[]
%     \centering
%     \includegraphics[width=0.75\textwidth]{figures/topL_ablation.pdf}
%     \caption{\textbf{Ablation study to evaluate the effect of neighbor selection strategies.} 
% The adaptive strategy denotes selecting the top-$L$ neighbors proportionally to the number of cluster centroids $K$ for each WSI (i.e., $L = 0.5K$). }
%     \label{fig:ablation_topL}
% \end{figure}

\begin{figure}[t]
    \centering
    \begin{minipage}[t]{0.48\textwidth}
        \centering
        \includegraphics[width=\textwidth]{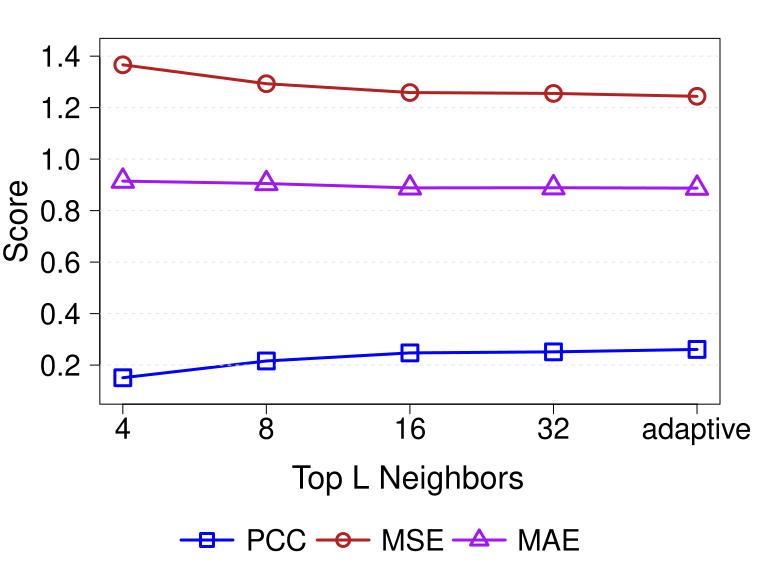}
        \caption{Impact of the number of the selected prototypes $L$ for a WSI. \texttt{adaptive} means $L = 0.5K$, where $K$ is the number of clusters corresponding to the given WSI.}
        \label{fig:ablation_topL}
    \end{minipage}%
    \hfill
    \begin{minipage}[t]{0.48\textwidth}
        \centering
        \includegraphics[width=\textwidth]{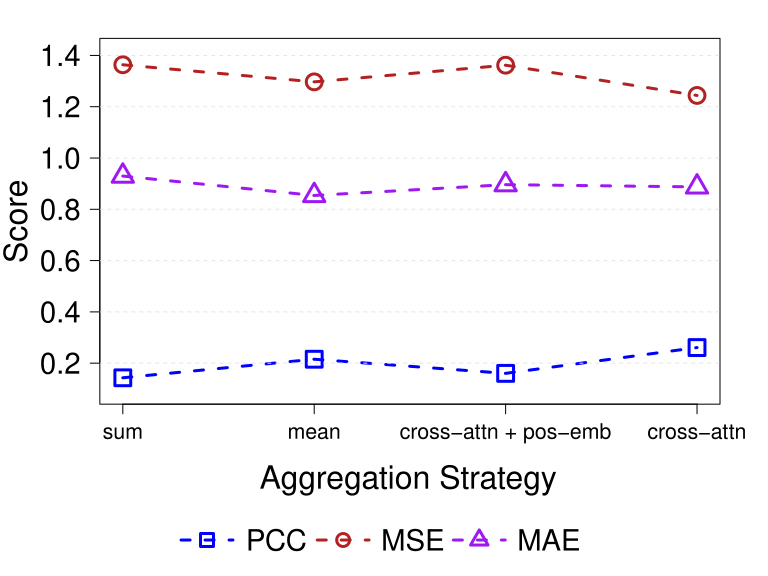}
        \captionsetup{font=footnotesize}
        \caption{\footnotesize Impact of four global context aggregation strategies:
        \texttt{mean} (element-wise mean), \texttt{sum} (element-wise sum), \texttt{cross-attn + pos-emb} (cross-attention with relative positional embeddings), and \texttt{cross-attn} (default cross-attention without positional encoding).}
        \label{fig:ablation_aggregation}
    \end{minipage}
\end{figure}

\subsubsection{Effect of Neighbor Selection Strategy.} We conduct an ablation study to evaluate the impact of neighbor selection strategies during the second phase, where each patch of a WSI attends to its top-$L$ most similar cluster centers. We compare fixed values of $L \in \{4, 8, 16, 32\}$ against an adaptive strategy, which selects the top 50\% of clusters (i.e. $L = 0.5K$, where $K$ is the number of clusters in the WSI) based on cosine similarity of the patch under consideration. As shown in Fig.~\ref{fig:ablation_topL}, the adaptive approach consistently achieves higher PCC and lower MSE, MAE. We observe that increasing $L$ generally leads to improved performance across all metrics, as a larger number of neighbors provide richer contextual information, helping the model better capture global structure and reduce noise in gene expression estimation. However, overly large or fixed values of $L$ may still include irrelevant or redundant context, especially in slides with highly variable spatial distributions. In contrast, the adaptive strategy we propose offers a principled way to balance this trade-off by dynamically adjusting the number of neighbors based on the number of clusters per WSI. This flexibility enables MMAP to better adapt to local tissue complexity, yielding more consistent improvements in PCC, MSE, and MAE.

% \begin{figure}[]
%     \centering
%     \includegraphics[width=0.75\textwidth]{figures/a}
%     \caption{\textbf{Ablation study on global context aggregation.} 
%     We compare different strategies for aggregating information from neighboring cluster centers in phase two of the model. 
%     The variants include: \texttt{mean} (element-wise mean over neighbor embeddings), \texttt{sum} (element-wise sum), \texttt{cross-attn + pos-emb} (cross-attention augmented with relative positional embeddings based on coordinate offsets), and \texttt{cross-attn} (the default cross-attention module without explicit positional encoding).}
%     \label{fig:ablation_aggregation}
% \end{figure}

\subsubsection{Ablation Study on Global Context Aggregation.}
We further investigate the role of the cross-attention module in aggregating information from neighboring cluster centers. Specifically, we replace the cross-attention block in phase two with three alternatives: (i) element-wise mean, (ii) element-wise sum over neighbor embeddings, and (iii) a variant that augments cross-attention with relative positional embeddings, computed as the coordinate offset between the center of the patch under consideration and the centers of its top-$L$ neighbors. As shown in Fig.~\ref{fig:ablation_aggregation}, the original cross-attention design consistently outperforms all alternatives across three evaluation metrics. This highlights its advantage in assigning adaptive relevance weights to neighboring regions based on learned contextual cues, rather than relying on fixed aggregation strategies. Notably, incorporating relative positional embeddings led to a consistent drop in performance. We attribute this to the possibility that explicit spatial encoding introduces inductive biases that overly constrain the attention mechanism. Since the cluster centers already encapsulate local structural information derived from spatially contiguous regions, enforcing additional position-based priors may hinder the model’s flexibility in capturing semantically relevant but spatially distant patterns, ultimately impairing generalization in gene expression prediction.

\section{Conclusion}
In this study, we propose MMAP, a powerful two-phase deep learning model to predict gene expression profiles from histology images by combining both local information and global context. MMAP demonstrates strong performance in predicting spatial gene expression from H\&E-stained histology images by effectively leveraging multi-magnification features and global context refinement. Unlike prior approaches that either overemphasize local features or rely heavily on computationally expensive global modeling, MMAP balances both through a prototype-based aggregation mechanism and a cross-magnification attention design. Our results show that MMAP consistently outperforms existing methods in prediction accuracy, confirming the utility of integrating hierarchical tissue information. These advantages allow MMAP to be more applicable in many real-world scenarios.

\begin{credits}
\subsubsection{\ackname} This study was funded by Hanoi University of Science and Technology (HUST) (Grant number T2024-TĐ-002)

\subsubsection{\discintname}
The authors declare that they have no known competing interests.

\subsubsection{Data availability.}
The spatial transcriptomics dataset that we used for this study can be found here at: \underline{https://github.com/almaan/her2st}

\subsubsection{Code availability.}
All source codes for baselines we used for comparison in our experiments are available at: \underline{https://github.com/lemonsoda174/MMAP-Baselines}. 
The source code for MMAP is available at: \underline{https://github.com/pndh/MMAPattemp3}. 
% (optional) acknowledgments\footnote{If EquinOCS, our proceedings submission
% system, is used, then the disclaimer can be provided directly in the system.},

\end{credits}
\bibliographystyle{splncs04}
\bibliography{references}
\end{document}